\documentclass[10pt,twocolumn,letterpaper]{article}

\usepackage{cvpr}
\usepackage{times}
\usepackage{epsfig}
\usepackage{graphicx}
\usepackage{amsmath}
\usepackage{amssymb}
\usepackage{comment}

\usepackage[breaklinks=true,bookmarks=false]{hyperref}

\cvprfinalcopy 

\def\cvprPaperID{****} 

\begin{document}

\title{3-D Reconstruction from Full-view Fisheye Camera}

\author{
	Chuiwen Ma\\
	Stanford University\\
	{\tt\small chuiwenm@stanford.edu}
	\and
	Liang Shi\\
	Stanford University\\
	{\tt\small liangs@stanford.edu}
	\and
	Hanlu Huang\\
	Stanford University\\
	{\tt\small hanluh@stanford.edu}
	\and
	Mengyuan Yan\\
	Stanford University\\
	{\tt\small mengyuan@stanford.edu}
}

\maketitle

\begin{abstract}
	In this report, we proposed a 3D reconstruction method for the full-view fisheye camera. The camera we used is Ricoh Theta, Fig. \ref{ricoh}, which captures spherical images and has a wide field of view (FOV). The conventional stereo apporach based on perspective camera model cannot be directly applied and instead we used a spherical camera model to depict the relation between 3D point and its corresponding observation in the image. We implemented a system that can reconstruct the 3D scene using captures from two or more cameras. A GUI is also created to allow users to control the view perspective and obtain a better intuition of how the scene is rebuilt. Experiments showed that our reconstruction results well preserved the structure of the scene in the real world.
\end{abstract}

\section{Introduction}

Wide field of view (fisheye) camera has received increasing attention over the past few years with its broad applications in surveillance, robotics, intelligent vehicles, immersive virtual environment construction, etc. For example, Nissan Motors developed a visual system that consists of four fisheye cameras mounted on the four sides of the vehicle. They together cover the entire 360$^\circ$ surrounding scene and allow drivers to examine all the visual blind spots that may cause danger. In surveillance, IP fisheye camera has become extremely prevalent for its wide cover range and easy axcessibility. Samsung provides a product with over 5 megpixel and 360$^\circ$ FOV, which is equipped in an alarm system performing intelligent motion detection, audio detection, and tampering detection. The supporting de-warping software allows users to undistort any subregion in the captured image.
\begin{figure}
	\centering
	\includegraphics[width = 2.5in]{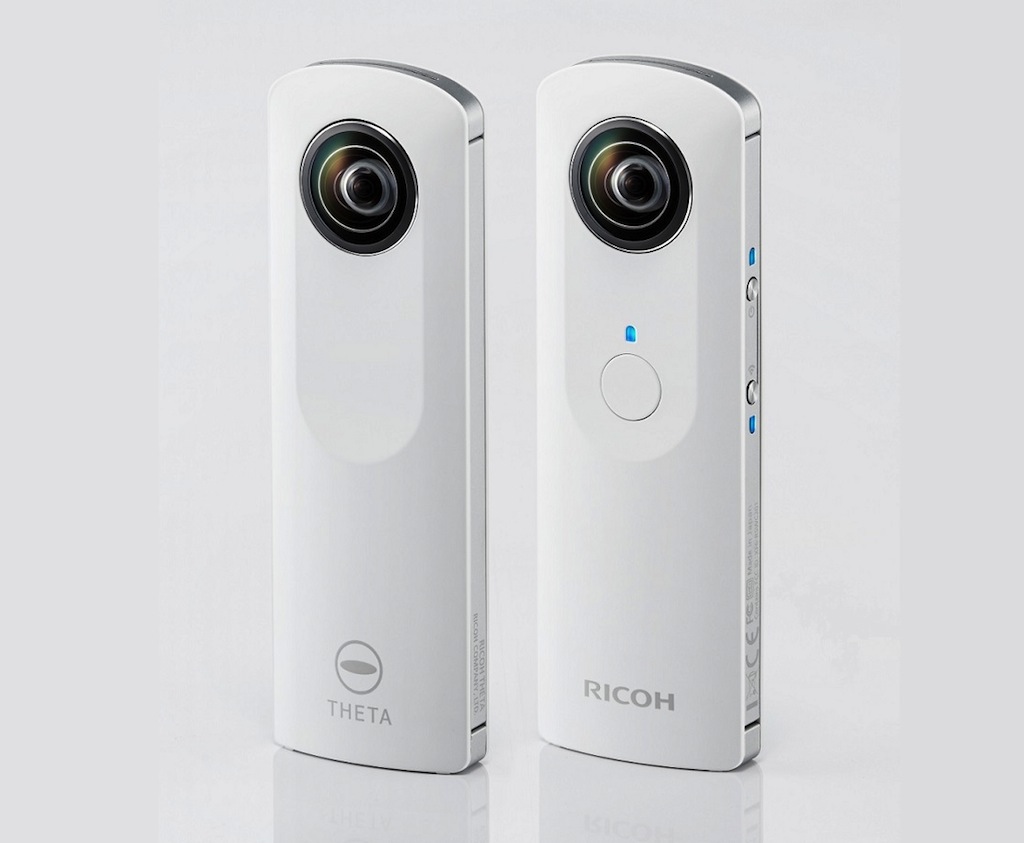}
	\caption{Ricoh Theta camera}
	\label{ricoh}
\end{figure}
\begin{figure}
	\centering
	\includegraphics[width = 2.5in]{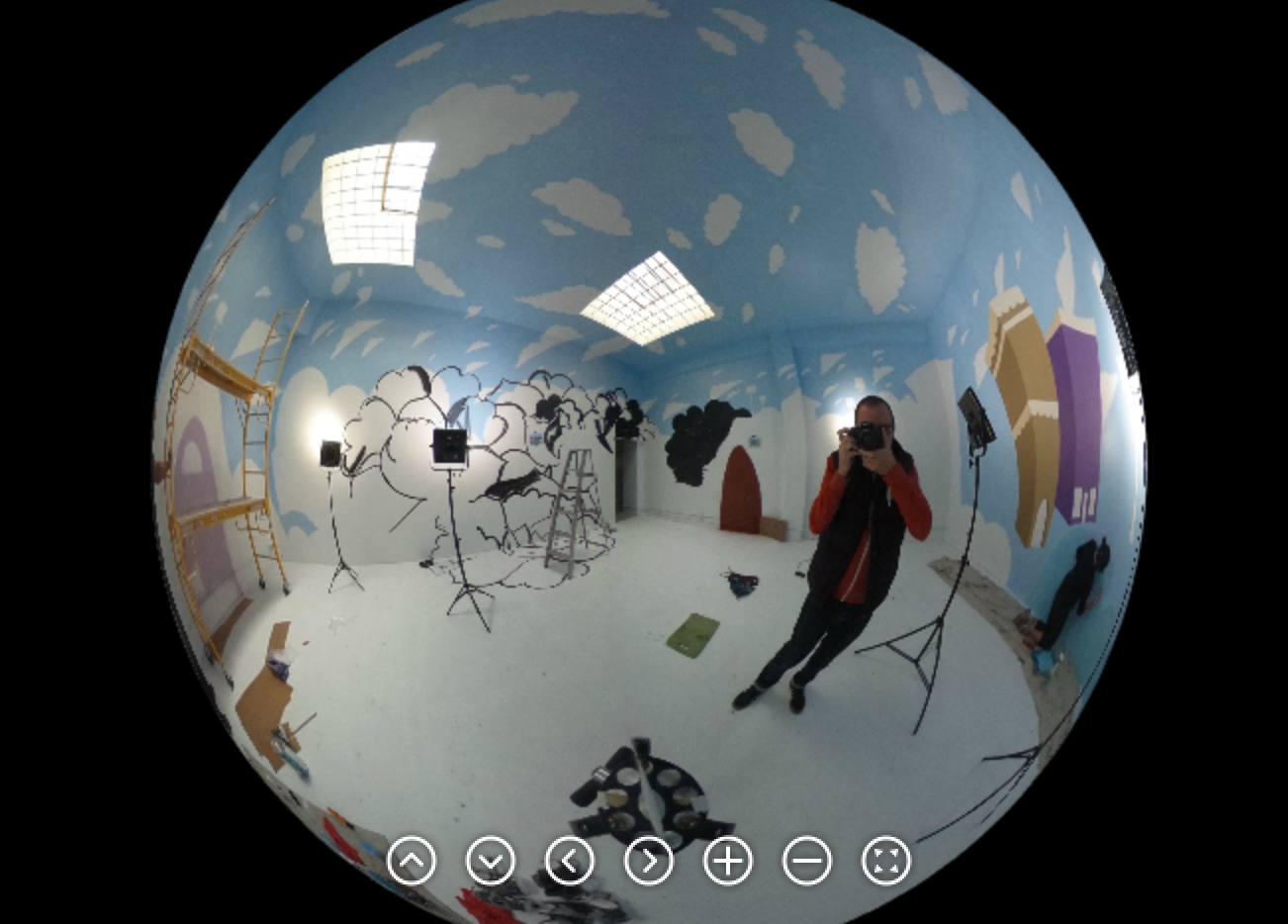}
	\caption{Spherical image captured by Ricoh Theta}
\end{figure}

Recently, Ricoh unveilled its first personal 360$^\circ$ fisheye camera --- Ricoh Theta. Two fisheye cameras are embedded on both front and back sides, to capture the entire scene with one click. Then the two captured images are stitched together to provide a dynamic 360$^\circ$ view with adjustable perspective controlled by the user. With this portable and handy device, our project aims to reconstruct the 3D scene using the captured spherical images. One important advantage of using this camera is that we are no longer required to set up multiple traditional cameras at different locations and directions to cover the entire scene. As a tradeoff, traditional camera model with perspective projection cannot be directly applied since fisheye camera has large radial distortion, especially near the border. To establish a one-to-one mapping between the 180$^\circ$ scene and a circular image, we created a model based on spherical projection. Based on this model, we can develop the epipolar geometry for fisheye cameras and solve the triangulation problem with least-square method. We used manually selected points to calculate the fundamental matrix, then applied it as a filter to prune the SIFT [\ref{Lowe}] matching result, at last augmented the point correspondences for reconstruction. On the other hand, we also tried dense reconstruction by first doing image rectification and then calculating the disparity map.

In Sec. \ref{RelatedWork}, we will briefly review previous work about 3D reconstruction with fisheye camera. Then in Sec. \ref{model} we will jump into the details of our camera model, revised epipolar geometry and data augmentation. Extension to multicamera registration and dense reconstruction will also be illustrated. In Sec. \ref{Experiment}, we first show the reconstruction result using hand-picked points, then we show the SIFT augmented result. Next, we give the disparity map and dense reconstruction result. Finally, we will show a snapshot of our GUI and provide the source code package for users to taste.

\section{Related Work}\label{RelatedWork}
\subsection{Previous work}
Perspective camera model is the most popular camera model in 3D reconstruction. However, it is limited for its narrow field of view. On the other hand, fisheye cameras which can capture spherical images have been paid more attention to during recent years. The major advantage is the wide FOV and thus more information it can incorporate from the environment.

Shah and Aggarwal [\ref{Shah}] presented an autonomous mobile robot navigation system in an indoor environment using two calibrated fisheye sensors. Micusik et al. [\ref{Micusik}] proposed a 3D reconstruction of the surrounding scene with two or more uncalibrated fisheye images. Li [\ref{Li}] drew 3D reconstruction by computing spherical disparity maps using binocular fisheye camera, which first calibrated the binocular camera to rectify the captured images and then used the correlation-based stereo to acquire the dense 3D representation of some simple environment. Herrera et al. [\ref{Herrera}] and Moreau et al. [\ref{Moreau}] placed the camera upwards and retrieved the environment information from the images. They computed disparity maps without image rectification step.

\subsection{Project contribution}
\begin{itemize}
	\item Proposed the camera model and epipolar geometry for fisheye camera.
	\item Designed a method to estimate camera rotation and position from point correspondences in multiple images.
	\item Implemented SIFT feature extraction and matching algorithm through equirectangular-to-cube mapping.
	\item Proposed sparse \& dense 3D reconstruction algorithm from multiple images.
	\item Developed a graphical user interface to interactively show multiple correlated 360$^\circ$ images.
\end{itemize}

\section{Method}\label{model}
In this section, we go over the mathematical model behind this project. It mainly consists of four parts, the fisheye camera model, epipolar geometry, multicamera registration and image rectification for dense reconstruction.
\subsection{Fisheye camera model}
\begin{figure}
	\centering
	\includegraphics[width = 1.2 in]{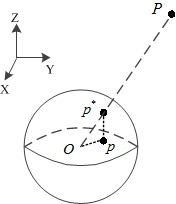}
	\caption{Fisheye camera model}
	\label{camModel}
\end{figure}
\begin{figure}
	\centering
	\includegraphics[width = 2.4in]{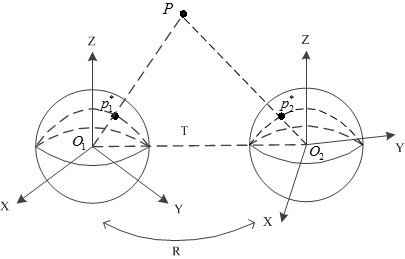}
	\caption{Epipolar geometry}
	\label{epipolar}
\end{figure}
The fisheye camera model is based on spherical projection. Suppose there is a sphere of radius $f_s$ and a point $P$ in space, as shown in Fig. \ref{camModel}. First, $P$ is projected to $p^*$ which is the intersection of the sphere surface with the line defined by sphere center $O$ and point $P$. This defines a mapping between spatial points to points on the sphere surface. Then, these points are vertically projected onto the image plane as $p^*$ is projected to $p$, which results in a circular image. In mathematical term, let $P = [X_p, Y_p ,Z_p]^T$, then we have $p^* = [f_s\sin\phi\cos\theta, f_s\sin\phi\sin\theta, f_s\cos\phi]^T$. The relation between $P$ and $p^*$ is
\[p^* = \lambda P\]
where $\lambda = f_s / \rho \ $ and $\rho = \sqrt{X_p^2 + Y_p^2 + Z_p^2}$.  The vertical projection reduces the $Z$ component to 0, and we get $p = [f_s\sin\phi\cos\theta, f_s\sin\phi\sin\theta, 0]^T$. Here, we let $f_s = 1$ which means we project onto a unit sphere.

The raw images acquired by Ricoh Theta, shown in Fig. \ref{fig:2view} are in equirectangular form with resolution 1024$\times$2048, i.e. the $(x,y)$ image coordinates represent the longitude and the latitude on the unit sphere.
\[\phi = y/1024*\pi\]
\[\theta = x/1024*\pi\]
\[p^*=[\sin\phi\cos\theta, \sin\phi\sin\theta, \cos\phi]^T\]

\subsection{Epipolar geometry}\label{EpiGeo}
Now, assume there are two cameras centered at $0$ and $T$, as shown in Fig. \ref{epipolar}. There is a point $P$ in 3D space. Then, for camera 1, the projection on spherical surface is $p_1^* = P/\left\|P\right\|$; for camera 2, the projection on spherical surface (in world coordinates) is $p_2^* = T+(P-T)/\left\|P-T\right\|$. Without loss of generality, assume the reference system of camera 1 is the same as the world reference system, and the rotation and translation between camera 1 and camera 2 is $R$ and $T$. Then, we have $z_{p,1} = Ip_1^*, z_{p,2} = R (p_2^*-T)$, where $z_{p,1}$ and $z_{p,2}$ are the coordinates of $p_1^*$ and $p_2^*$ in their cameras' reference system.

Notice, now we have five coplanar points: camera centers $O_1,O_2$, $p_1^*,p_2^*$ and $P$. Thus, we have the constraint $\overrightarrow {{O_1}p_1^*} \cdot (\overrightarrow {O_1O_2}\times \overrightarrow {O_2 p_2^*}) = 0$. I.e. $(p_1^*)^T (T\times (p_2^*-T))=0$. Substitute $p_1^*, \ p_2^*$ with $z_{p,1}$ and $z_{p,2}$, we get,
\[z_{p,1}^T \cdot [T_\times]\cdot R^{-1} z_{p,2} = 0\]
Define $F=[T_\times]\cdot R^{-1}$ as the fundamental matrix for fisheye camera pair, we have constraint $z_{p,1}^T F z_{p,2}=0$. Now, we can use the eight-points algorithm or RANSAC to solve for $F$. Once we get the fundamental matrix, we can calculate the epipoles in the two cameras by solving $F^T e_1=0$, $Fe_2=0$.

Recall that the definition of epipoles is $e_1 = T/\left\|T\right\|$, $e_2 = -RT/\left\|T\right\|$, which gives $e_2 = -Re_1$. 
Then the rotation matrix $R$ can be derived as,\\
\[R = I + [v]_{\times} + [v]_{\times}^2\frac{1-c}{s^2}\]
where $v = (-e_1)\times e_2$, $s = \left\|v\right\|$, $c = -e_1^T e_2$. Here we assume the Euclidean distance between $O_1$ and $O_2$ is 1, i.e. $T = e_1$.

Now, we can triangulate $P$ using parameters $z_{p,1}$, $z_{p,2}$, $e_1$, and $R$. We define the line passing through $O_1$ and $z_{p,1}$ as $az_{p,1}$, where $a\in\mathbb{R}$; the line passing through $O_2$ and $z_{p,2}$ as $b R^{-1}z_{p,2} + e_1$, $b\in\mathbb{R}$. The goal of triangulation is to find the minimal distance between the two lines. We can formulate this into a least square problem,\\
\[\begin{array}{*{20}{l}}
{{\rm minimize}_{a,b}}&{ \left\| az_{p,1} - b R^{-1} z_{p,2} -e_1\right\|}
\end{array}\]
where the optimal solution is given by,
\[\left[ {\begin{array}{*{20}{c}}
	{{a^\star}}\\
	{{b^\star}}
	\end{array}} \right] = {\left( {{A^T}A} \right)^{ - 1}}{A^T}{e_1},\quad A = \left[ {\begin{array}{*{20}{c}}
	{{z_{p,1}}}&{ - {R^{ - 1}}{z_{p,2}}}
	\end{array}} \right]\]
Once we get the optimal parameter $a^\star$, $b^\star$. The minimal distance is known to be achieved between $a^\star z_{p,1}$ and $b^\star R^{-1} z_{p,2} +e_1$, then $P$ can be assigned as the their middle points: 
\[P = (a^\star z_{p,1}+b^\star R^{-1} z_{p,2} +e_1)/2\]

\subsection{$F$ estimation \& data augmentation}
In order to calculate $F$, we must have enough point correspondences in multiple images. In our methods, we manually selected around 45 pairs of corresponding points. We also attempted to automatically estimate $F$ by applying RANSAC with constraint $z_{p,1}^TFz_{p,2} = 0$ on SIFT  matching results, to find the best estimation of $F$. However, SIFT matching is not invariant to radial distortion and the matching results have unacceptable outliers, thus the estimation of $F$ is not robust enough. Instead, we estimated $F$ by using hand-picked points, and in turn use $F$ to filter the SIFT matches and extend our point pairs pool. 

\subsection{Multicamera registration}
Next, we extend the discussion to multi-view scenario. From the section above we can obtain the fundamental matrix and epipoles for each pair of cameras, but we can no longer assume the Euclidean distance between camera centers is 1. Now we want to estimate the rotation matrix and camera position for each camera. This can be done in a two-step process.

First, we estimate the rotation for each camera. Assume we have $n$ cameras, for each pair of camera $i$ and $j$, we can calculate the epipoles $e_{ij}$ and $e_{ji}$, which denotes $O_j$ on image $i$ and $O_i$ on image $j$, respectively. Here, we assume the cameras all lie on the same horizonal plane, which is a very good approximation of how we took pictures. Therefore, the rotation of each camera can be represented by an angle $\theta_i$. The relation between $\theta_i$ and rotation matrix $R_i$ is
\[
R_i = \left[ {\begin{array}{*{20}{c}}
	{ \cos(\theta_i)}&{-\sin(\theta_i)}&{0}\\
	{ \sin(\theta_i)}&{ \cos(\theta_i)}&{0}\\
	{0}&{ 0}&{ 1}
	\end{array}} \right]
\]

The epipole direction in world coordinate is
\[
e_{ij,w}=R_i^{-1}e_{ij}\]
\[
e_{ji,w}=R_j^{-1}e_{ji}\]
\[
e_{ij,w}=\pm e_{ji,w} \]

The last line should be obvious as they both denote the direction of the line segment defined by $O_i$ and $O_j$. The $\pm$ sign indicates the two-fold ambiguity in calculating $e_{ij}$ from the fundamental matrix.

Now we need to minimize the objective function,
\[
\sum_{i,j} 1-|e_{ij,w}\cdot e_{ji,w}|
\]
which is a convex optimization problem, and we solve it by Newton's method.

Next, we estimate the position of each camera. As we have the direction of each line segment $\overrightarrow {O_iO_j}$, this is a triangulation problem. A naive way to solve this problem is to choose two cameras, e.g. $O_1$ and $O_2$, and set the Euclidean distance between them to be 1. Then for each camera other than $O_1$ and $O_2$, its position can be triangulated from the direction of $\overrightarrow {O_1O_i}$ and $\overrightarrow {O_2O_j}$. We can repeat the procedure with different choice of baseline to check the consistency. We can also feed the result into another gradient descent program to adjust the camera positions using all directions $O_iO_j$ obtained.

Now that we have recovered the rotation and translation of each camera, the object points can be triangulated in a similar way as described in Sec. \ref{EpiGeo}. In the multiview case, we assign $r_i$ as the distance between object point and camera center $O_i$, and minimize the mean squared distance among the $n$ points obtained from each camera image.
\[
P_i=r_iR^{-1}z_{p,i}+P_{c,i}, \; i=1,2,...,n\]
\[P_{\rm mean}=\frac{1}{n}\sum_{i} P_i\]
\[\text{minimize } \sum_{i} \|P_i-P_{\rm mean}\|
\]

\subsection{Image rectification \& dense reconstruction}
\begin{figure}
	\centering
	\includegraphics[width = 5in]{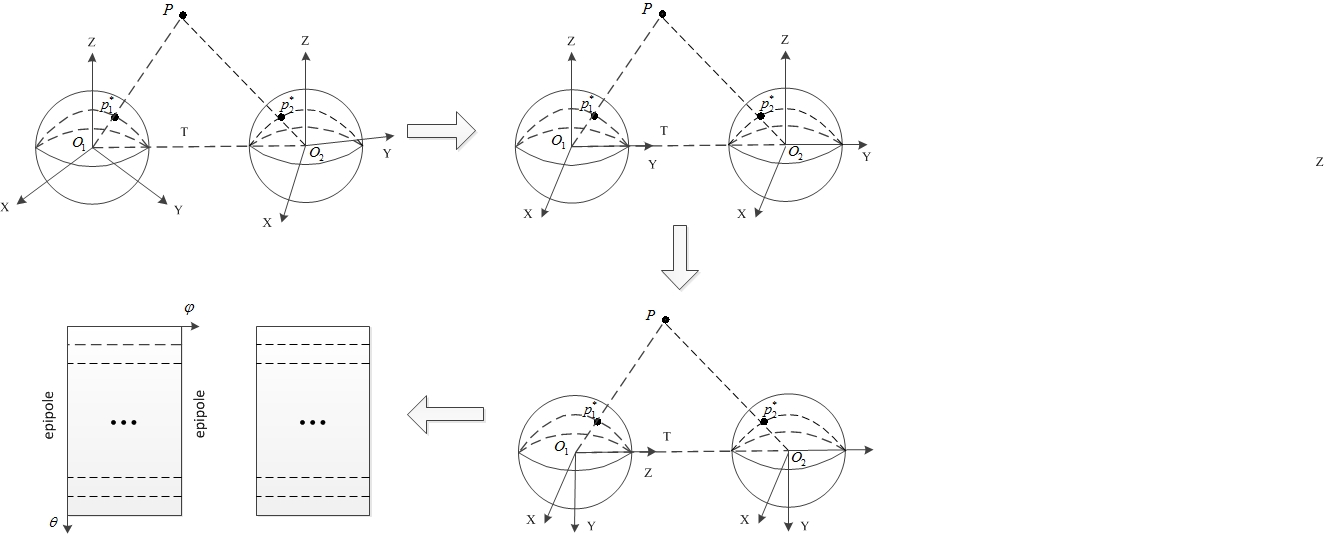}
	\caption{Image rectification pipeline}
	\label{rectPipe}
\end{figure}
Now, we want to go one step further from sparse reconstruction to dense reconstruction. In order to achieve a dense reconstruction, we need to rectify the image pairs so that their epipolar lines are horizontal and all corresponding points have the same vertical coordinate on the image. As we know, the epipolar lines in spherical images are circles which intersect with the epipoles. Therefore, if we rotate the camera reference such that the $Z$ axis align with the epipole and the $X$,$Y$ axis are parallel, and map the sphere onto equirectangular image, then epipolar lines would be vertical lines in equirectangular images, as show in Fig. \ref{rectPipe}. By exchanging the horizontal and vertical coordinates, the image pairs will be rectified. From the rectified image pairs we can calculate a disparity map $D$, which is the distance (in pixels) between corresponding points on the image pair. As the images are equirectangular, $D$ is the angle $\angle O_1PO_2$ by a constant. Assume the corresponding points are $(x_1,y)$ and $(x_2,y)$ respectively, and $d = x_2-x_1$, then $\|O_1P\|=T\times \sin(x_2)/\sin(d)$, $\|O_2P\|=T\times \sin(x_1)/\sin(d)$, where $T=\|O_1O_2\|$. From that we can calculate the 3D coordinates of point $P$.

\section{Experiment}\label{Experiment}
\subsection{Two cameras reconstruction}
\begin{figure}[h]
	\centering
	\includegraphics[width=0.48\textwidth]{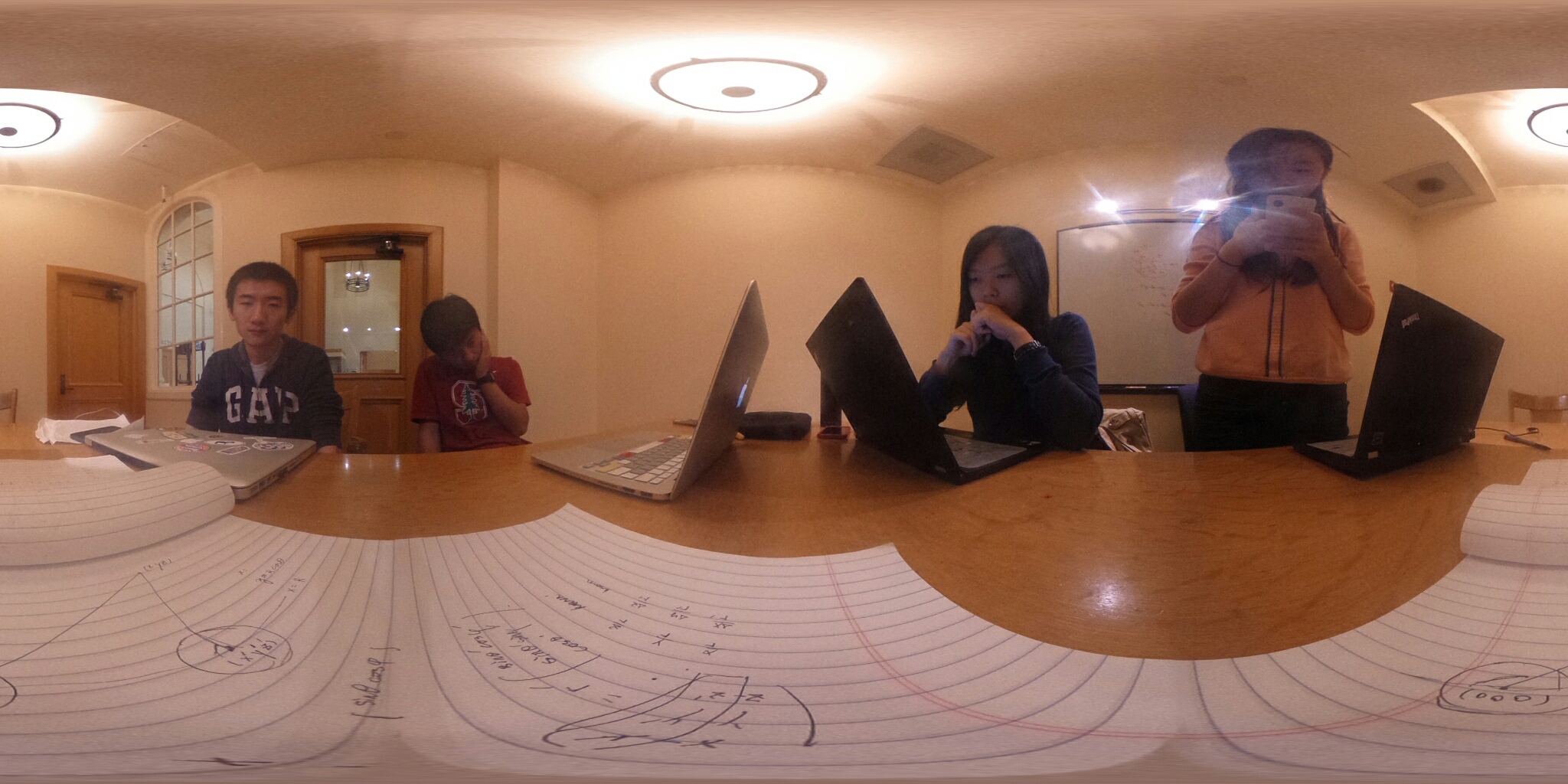}
	\includegraphics[width=0.48\textwidth]{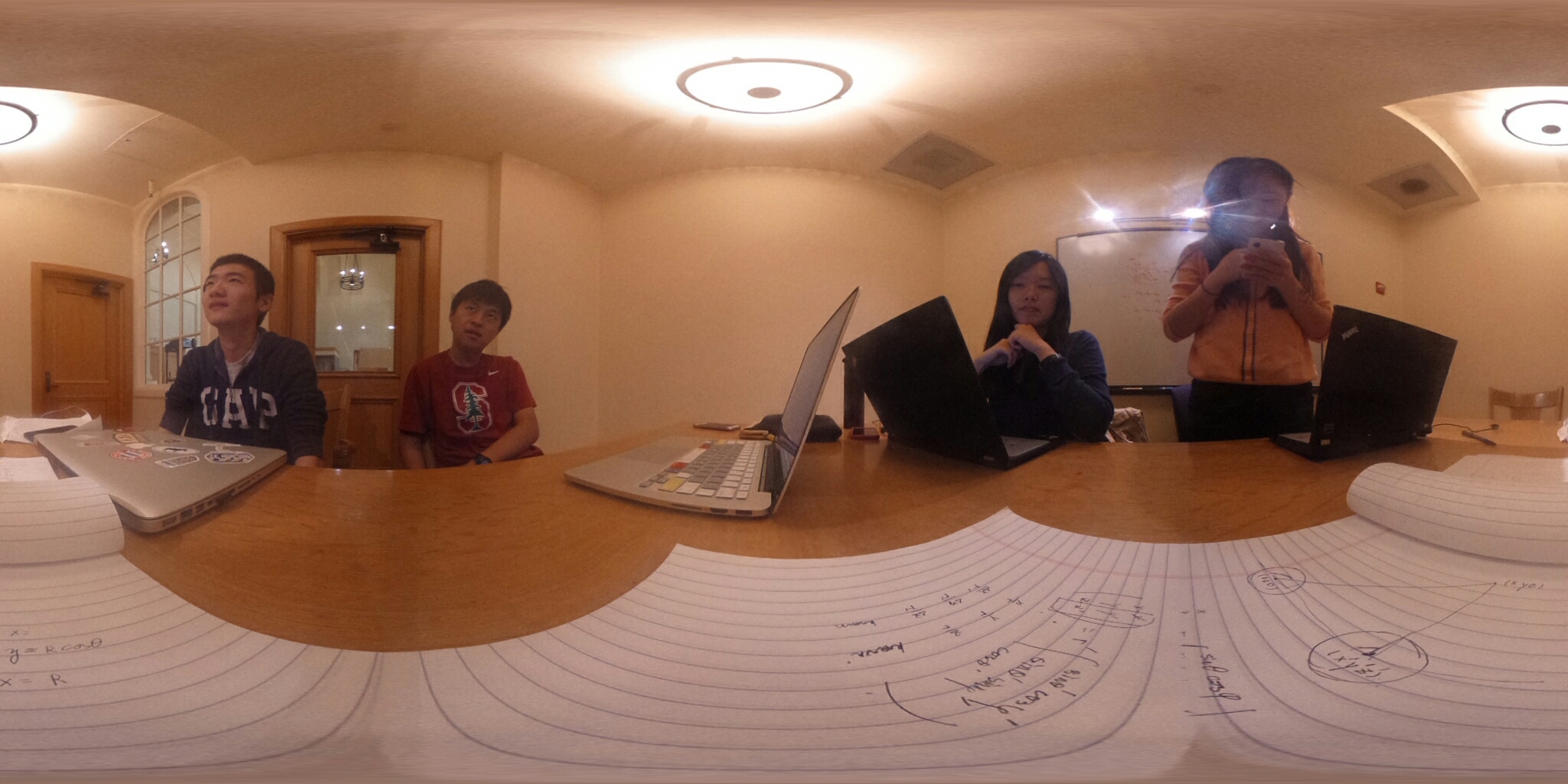}
	\caption{2-camera raw image}
	\label{fig:2view}
\end{figure}

In this section, we show the reconstruction results for a 2-cameras settings. The two raw images are shown in Fig. \ref{fig:2view}.

\subsubsection{Ground truth point matching}
\begin{figure}[h]
	\centering
	\includegraphics[width=0.236\textwidth]{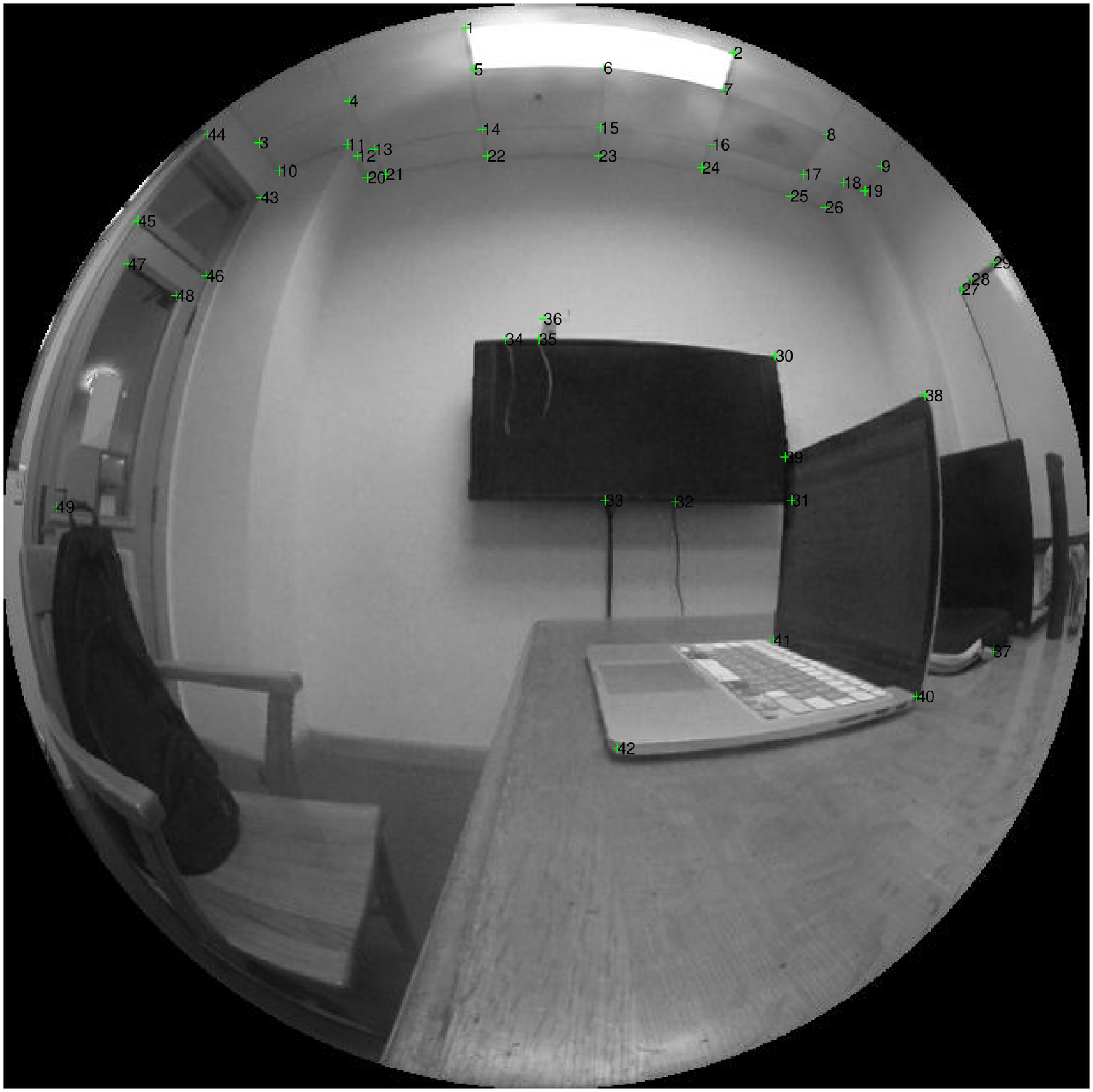}
	\includegraphics[width=0.236\textwidth]{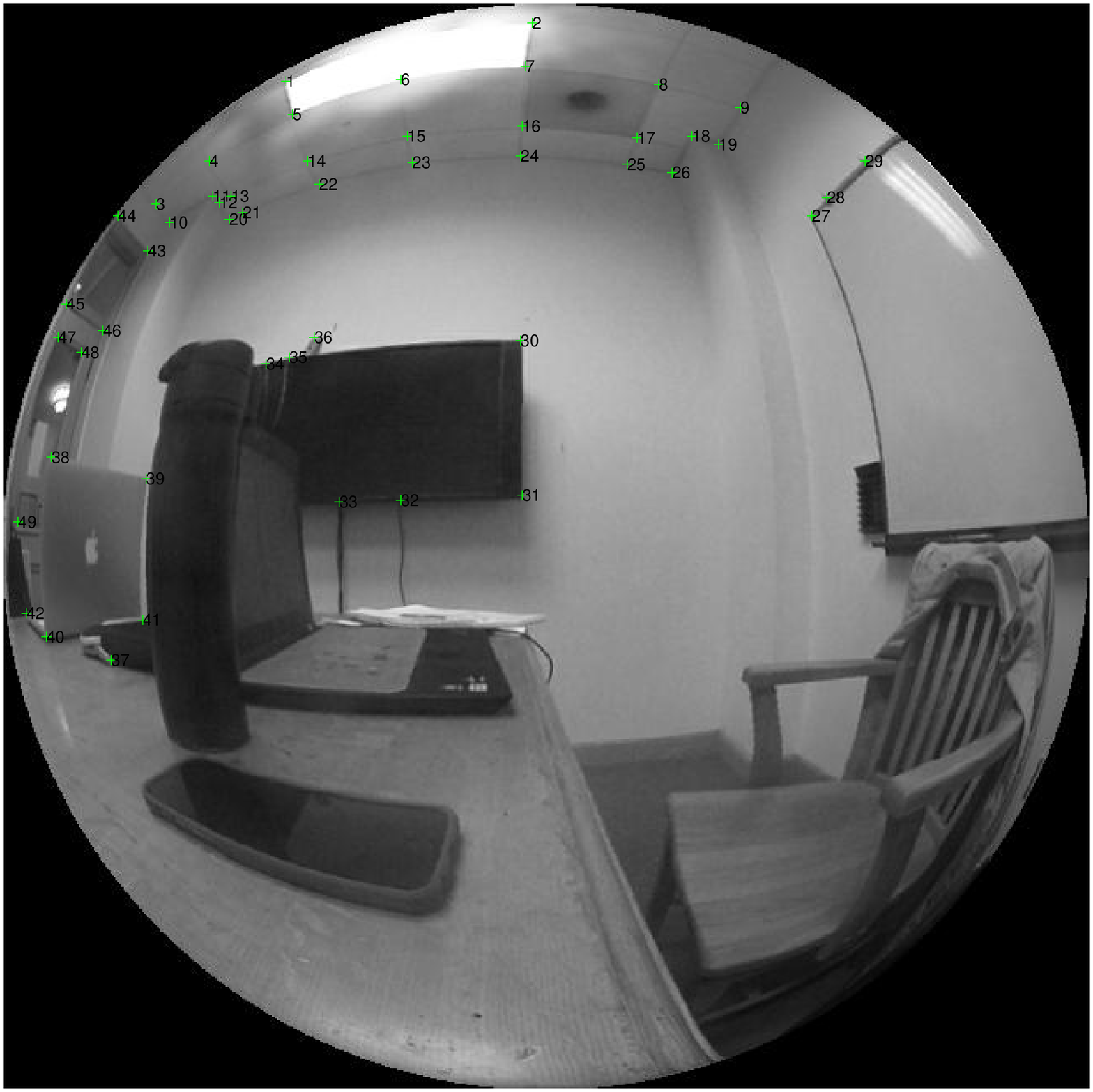}
	\caption{Ground truth point correspondences}
	\label{fig:ptsCorr}
\end{figure}

In order to implement the eight-point algorithm to compute the fundamental matrix, we manually labeled ground truth point correspondences on circular images, shown in Fig. \ref{fig:ptsCorr}. For each view, we labeled around 45 pairs of corresponding points, which are typically on the ceiling or on the walls, thus easy to recognize. There are also several points around the desk, such as the corner of the computer.
\subsubsection{SIFT point matcing}
\begin{figure}[h]
	\centering
	\includegraphics[height=0.49\textwidth]{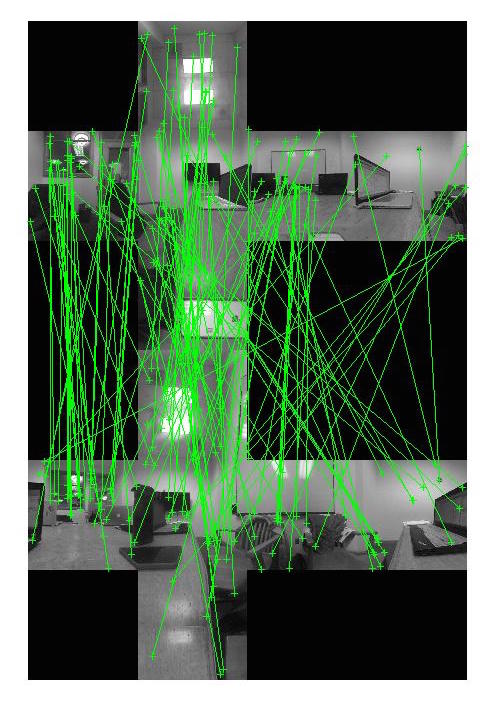}
	\caption{SIFT matching result on cubic images}
	\label{cubeSIFT}
\end{figure}

We also tried to extract point correspondences using SIFT, then estimate $F$ automatically using RANSAC. However, due to the large amount of outliers, this approach is not robust enough. Therefore, we proposed another pipeline --- use the ground truth $F$ to filter SIFT matching results, and add those correspondences into our correspondences pool to achieve a denser reconstruction result. In our project, we used the SIFT implementation in VLFeat toolbox [\ref{VLFeat}] for extracting features and performing point matching. We applied point mathcing on both raw images and cubic images achieved by cube mapping [\ref{Greene}]. Fig. \ref{cubeSIFT} shows a rough matching result using cubic images.

\subsubsection{3D reconstruction}
\begin{figure}[h]
	\centering
	\includegraphics[width = 0.49\textwidth]{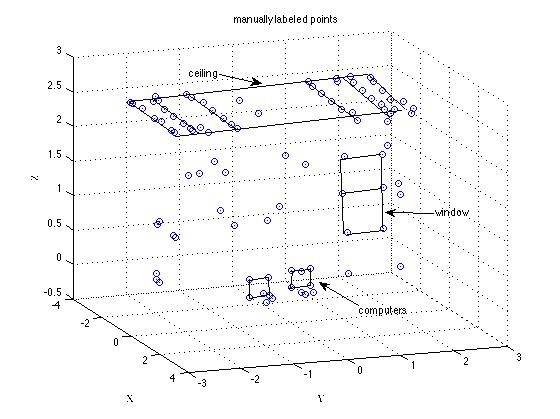}
	\includegraphics[width = 0.49\textwidth]{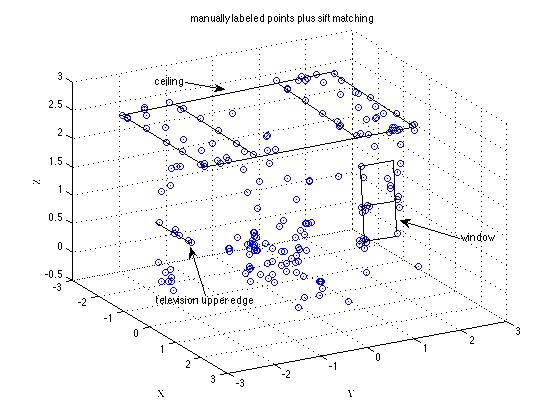}
	\caption{2-camera reconstruction result: the first image is the reconstruction result using only hand-picked points, the second is the reconstruction result augmented by SIFT.}
	\label{fig:3dRec}
\end{figure}

Using the ground truth fundamental matrix $F$, we calculated $e_1,e_2$ and $R$. Then, we used the triangulation method we proposed in Sec. \ref{EpiGeo} to recover points' position in 3D space. The reconstruction result is shown in Fig. \ref{fig:3dRec}.

\subsection{Multiple cameras reconstruction}
\begin{figure}[h]
	\centering
	\includegraphics[width = 3in]{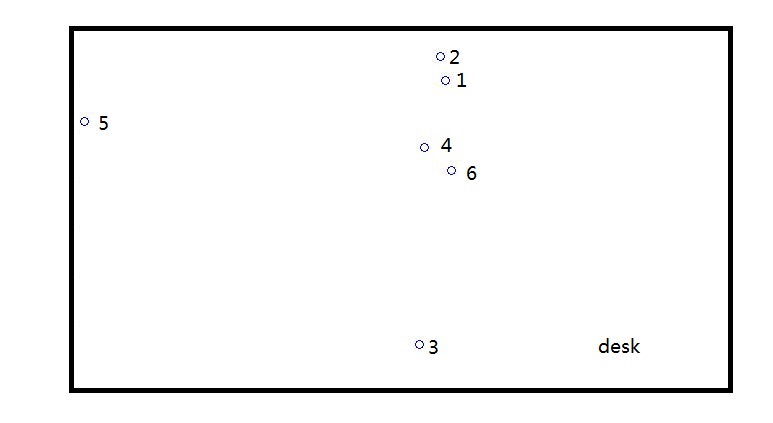}
	\caption{Camera positions in multiple reconstruction}
	\label{camPos}
\end{figure}

\begin{figure}[h]
	\centering
	\includegraphics[width = 3in]{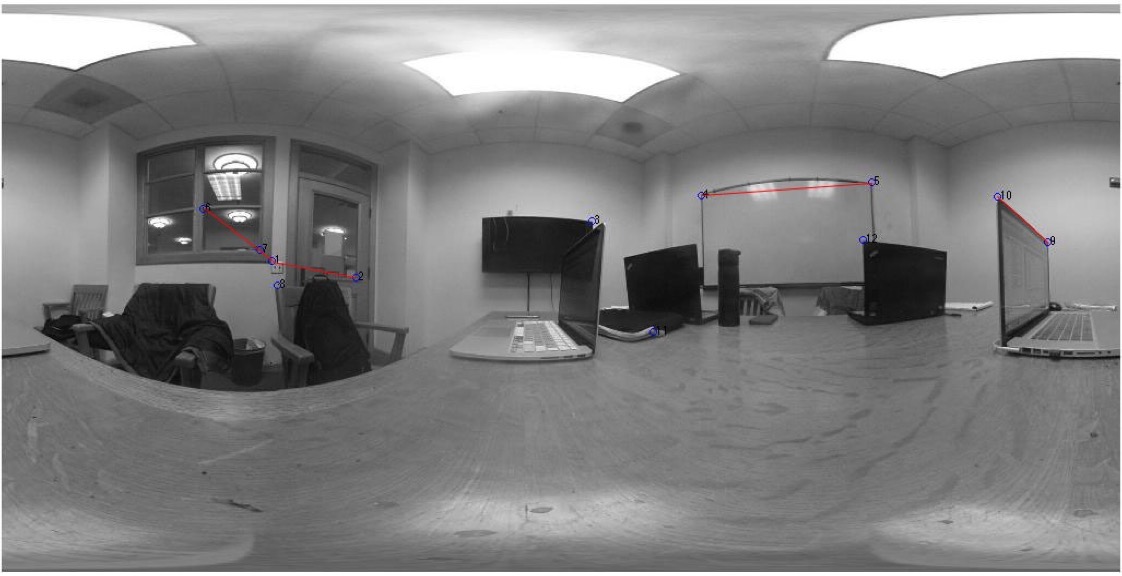}
	\includegraphics[width = 3in]{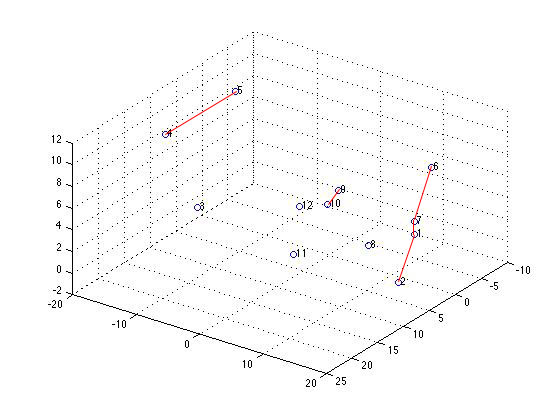}
	\caption{Sample points illustration \& results }
	\label{6cam}
\end{figure}

Now, we show the reconstruction result using pictures captured at 6 different loacations, which is equivalent to 6 cameras. We manually select 12 corresponding points on each of the 6 images. For each pair of image we calculate the fundamental matrix $F$ and epipoles $e_1$, $e_2$. Then, we calculated the rotation and position of each camera. The position of the cameras is shown in Fig. \ref{camPos}. Once the rotation and position of each camera is obtained, we can triangulate the corresponding points as well as rectify each pair of images. The 3D reconstruction for the 12 points is shown in Fig. \ref{6cam}. The result matches well with the ground truth.

\subsection{Disparity map \& dense reconstruction}
\begin{figure}[h]
	\centering
	\includegraphics[width = 0.153\textwidth]{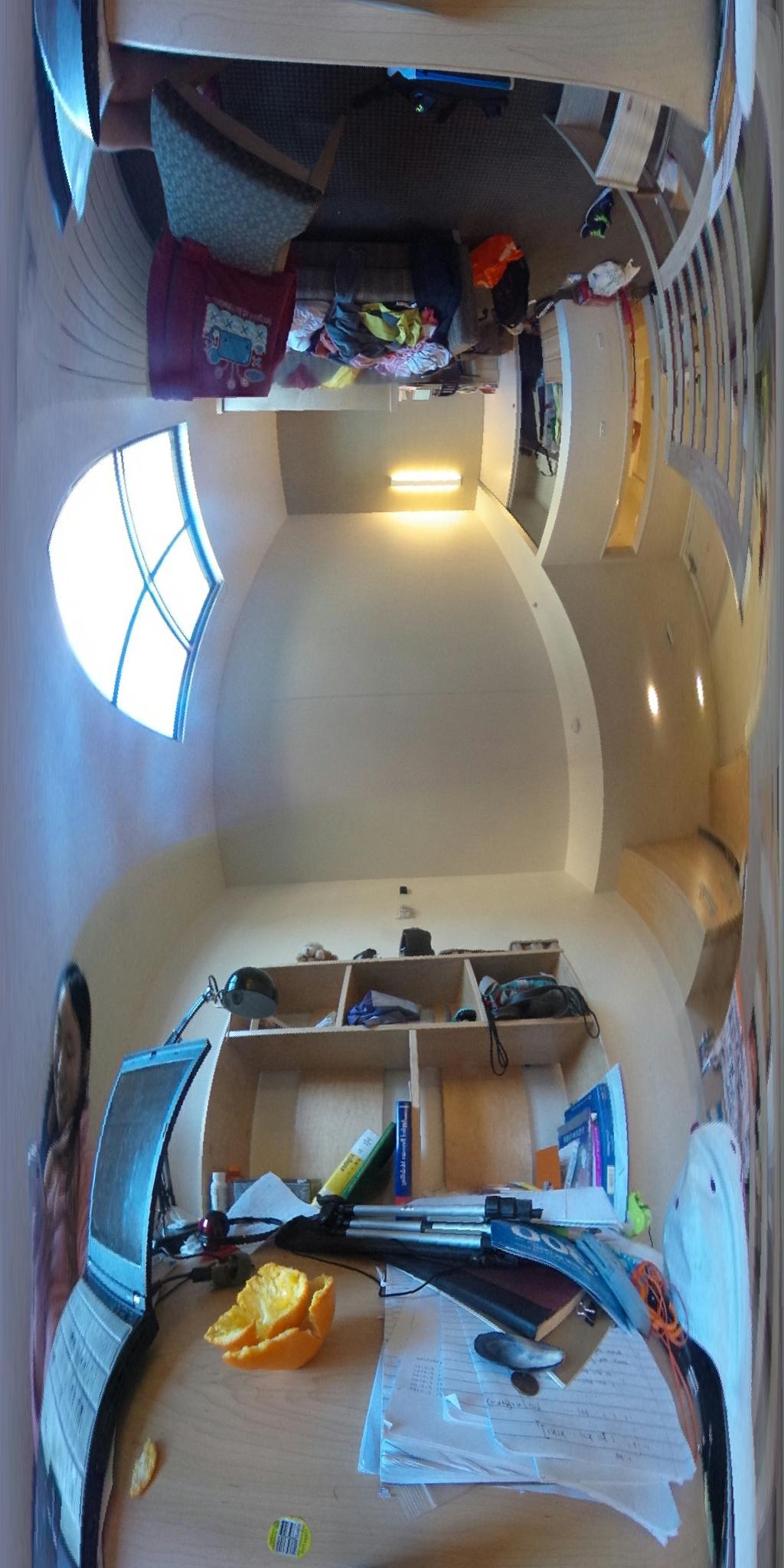}
	\includegraphics[width = 0.153\textwidth]{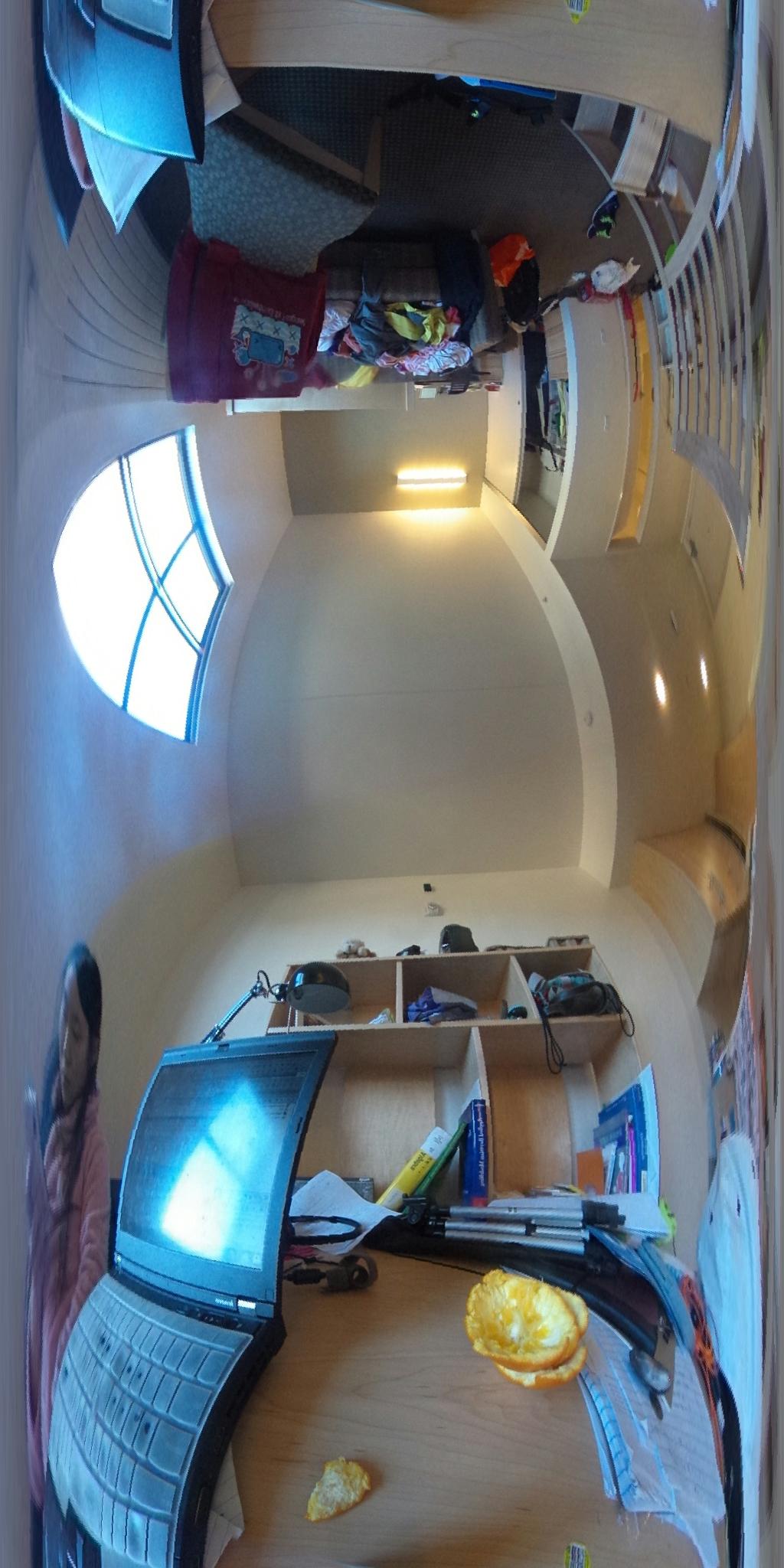}
	\includegraphics[width=0.153\textwidth]{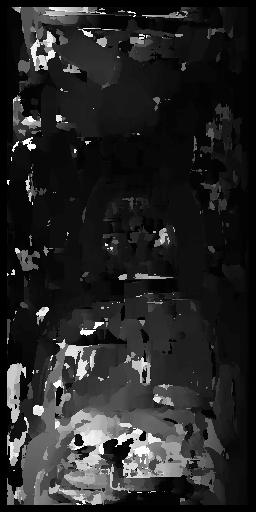}
	\caption{Rectified images \& disparity map}
	\label{Rectified}
\end{figure}

\begin{figure}[h]
	\centering
	\includegraphics[width=0.49\textwidth]{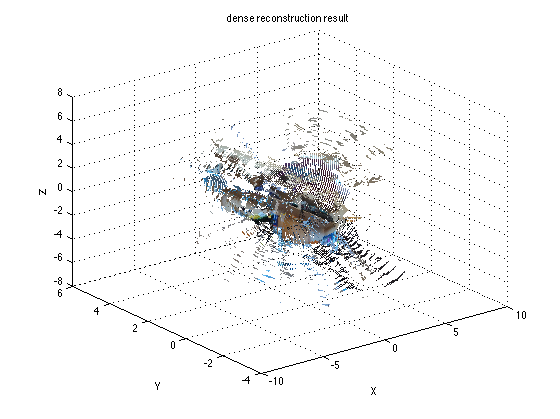}
	\caption{Dense reconstruction result}
	\label{denseRec}
\end{figure}

After rectification, the corresponding points are at the same longitude with each other. So after transforming the raw image into longitude-latitude image, we can use the traditional method to find the corresponding pairs in the images. The calculated disparity map is shown in Fig. \ref{Rectified}, together with the two rectified images. The brighter part means smaller disparity and the darker part indicates larger disparity. As we can see, the image have roughly presented the deapth information. While since the rectified images still have distortion, the disparity map may have noise. The reconstruction result is shown in Fig. \ref{denseRec}. Although the result looks a little messy, we can see the closet are reconstructed fairly well.

\subsection{GUI implementation}
\begin{figure}[h]
	\centering
	\includegraphics[width = 3in]{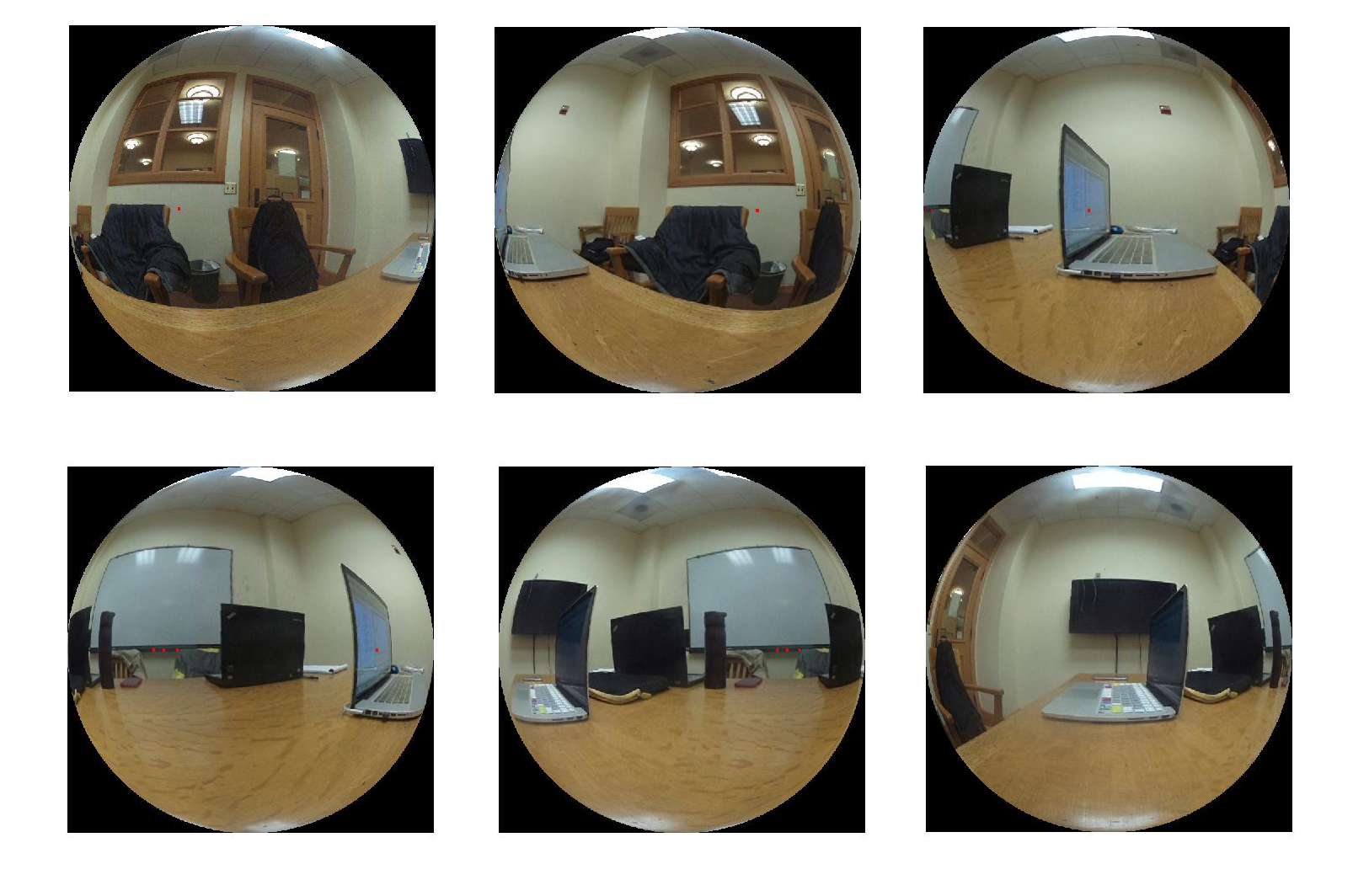}
	\caption{GUI demo}
	\label{GUI}
\end{figure}

A graphical user interface (GUI) is developed using the 6-view dataset. You can run \texttt{demos.m} to see the demonstration. 
Fig. \ref{GUI} gives a brief illustration of the 6 views obtained by user control.

\section{Conclusion}
In this project we implemented 3D reconstruction algorithm for multiple spherical images. We obtained our data using Ricoh Theta fullview fisheye camera. We used both manually selected points and SIFT matching points to estimate fundamental matrix for each pair of images. Then, we calculated epipoles, the rotation and the position of each camera. Based on these information we implemented sparse 3D reconstruction, the result matches well with the ground truth. We also developed a user interface to enable users to interactively view multiple correlated 360$^\circ$ images. Our project is an important step towards building virtual tour from large number of fullview images.

\section{Future Work}
There are two things we want to improve in the future. The first is to enhance the algorithm of generating disparity map. The second is the robustness of SIFT matching in various datasets. Currently the performance of SIFT matching fluctuates between different image sets. In outdoor images, SIFT matching performance tends to deteriorate, the reason could be that camera centers are too far apart thus image pairs differ too much, or that buildings tend to have repetitive features like arches, windows, etc. We could improve the image capturing behaviours and select more appropriate scenes to get a better performance.

\bibliographystyle{abbrv}

\end{document}